\relax
\documentclass[letterpaper]{article} 
\usepackage{aaai21}  
\usepackage{times}  
\usepackage{helvet} 
\usepackage{courier}  
\usepackage[hyphens]{url}  
\usepackage{graphicx}  
\usepackage{subfigure}
\usepackage{amsmath}
\usepackage{bm}
\usepackage{multirow}
\usepackage{amssymb}
\usepackage{booktabs}
\usepackage{graphicx}  
\usepackage{natbib}  
\usepackage{caption} 
\title{Few-shot Learning with LSSVM Base Learner and Transductive Modules}
\author{
    Haoqing Wang,
    Zhi-Hong Deng\\
}
\affiliations{

    Key Laboratory of Machine Perception (Ministry of Education), \\
    School of Electronics Engineering and Computer Science, Peking University, Beijing 100871, China \\
    wanghaoqing@pku.edu.cn, zhdeng@cis.pku.edu.cn\\
}
\begin{document}
\maketitle
\begin{abstract}
The performance of meta-learning approaches for few-shot learning generally depends on three aspects: features suitable for comparison, the classifier ( base learner ) suitable for low-data scenarios, and valuable information from the samples to classify. In this work, we make improvements for the last two aspects: 1) although there are many effective base learners, there is a trade-off between generalization performance and computational overhead, so we introduce multi-class least squares support vector machine as our base learner which obtains better generation than existing ones with less computational overhead; 2) further, in order to utilize the information from the query samples, we propose two simple and effective transductive modules which modify the support set using the query samples, i.e., adjusting the support samples basing on the attention mechanism and adding the prototypes of the query set with pseudo labels to the support set as the pseudo support samples. These two modules significantly improve the few-shot classification accuracy, especially for the difficult 1-shot setting. Our model, denoted as FSLSTM (Few-Shot learning with LSsvm base learner and Transductive Modules), achieves state-of-the-art performance on \emph{miniImageNet} and \emph{CIFAR-FS} few-shot learning benchmarks.
\end{abstract}

\begin{figure*}
\begin{center}
\includegraphics[width=0.95\linewidth]{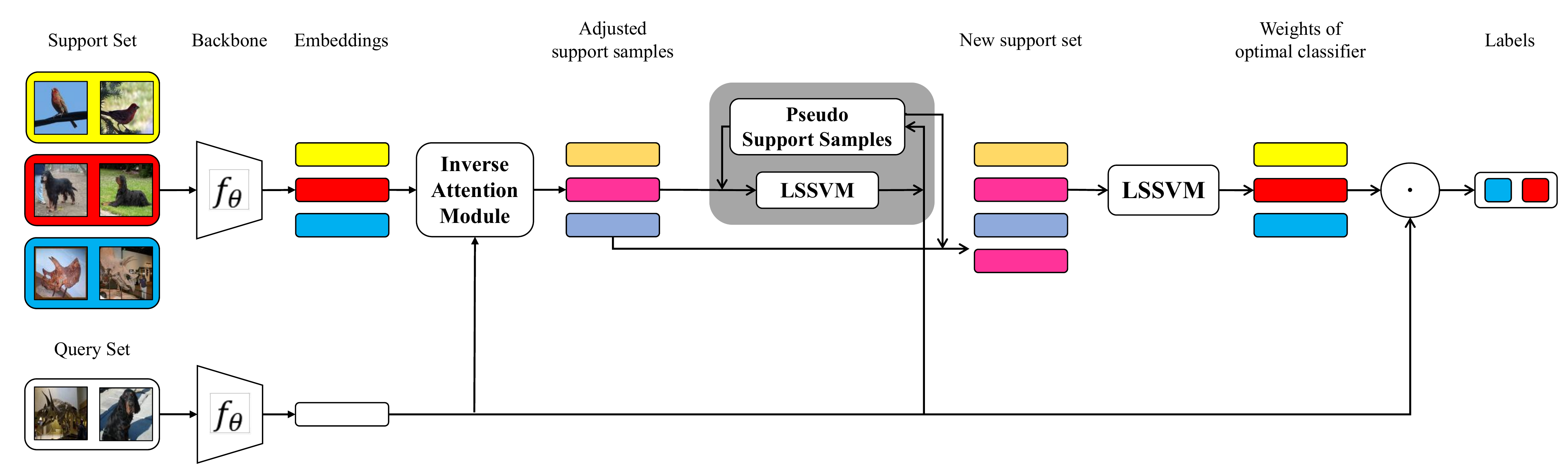}
\end{center}
\caption{\textbf{Overview of our model.} A few-shot classification task is the tuple of a support set and a query set, and we need to correctly classify the samples in the query set. In order to obtain a better classifier, we introduce LSSVM as our base learner (see Section \ref{LS_SVM}); and then we use the query samples to adjust the support samples basing on the attention mechanism (see Section \ref{TML}), corresponding to \textbf{Inverse Attention Module}; and we add the prototypes of the query set with pseudo labels to the support set as the new support samples (see Section \ref{TML}), corresponding to \textbf{Pseudo Support Samples}, and this operation can be iterated multiple times and is only used during meta-testing.}
\label{model}
\end{figure*}

\section{Introduction}

Human can learn from a few examples, yet it is a challenge for modern machine learning systems. Although many deep learning based image recognition approaches have achieved impressive performance \cite{simonyan2014very,he2016deep}, they are data-hungry which require hundreds of training samples from each class. Few-shot classification \cite{fei2006one,lake2015human} aims to classify unseen instances into a set of new classes based on a few support samples from each class and is more challenging.

Few-shot classification has received significant attention from the machine learning community recently and the research has been dominant by meta-learning based approaches. The performance of many approaches generally depends on three aspects: features suitable for comparison, the base learner suitable for low-data scenarios, and valuable information from the samples to classify. First, features satisfying the \emph{clustering assumption} are easier to be classified, i.e., there is a clustering structure and the samples in the same cluster belong to the same class, which inspires us to learn task-specific representations \cite{li2019finding,ye2020few} or improve the generalization ability of the backbone \cite{gidaris2019boosting,seo2020self}. Second, different classifiers, e.g., nearest-neighbor classifier, ridge-regression and SVM, have different classification capabilities in the low-data scenarios and computational overhead, which inspires us to introduce different classifiers as the base learner \cite{snell2017prototypical,bertinetto2018meta,lee2019meta}. Finally, considering the limited supervision information in the supporting samples in a few-shot task, the knowledge in the query samples is very valuable, which inspires us to perform transductive learning \cite{yang2020dpgn,hu2020empirical}. In this work, we focus on the last two aspects and make improvements. Our model is illustrated in Figure \ref{model}.

While many choices of base learners exist, there is a trade-off between generalization and computational overhead. SVM base learner \cite{lee2019meta} can achieve better generalization but with an increase in computational overhead, as the objective is a quadratic program problem and is solved with iterative algorithm. In this work, we propose to use multi-class least squares support vector machine (LSSVM) as the base learner which improves generalization with less computational overhead. Further, in order to utilize the information from the query samples to enhance classification, we propose two transductive modules and the main motivation is to use the query samples to modify the support set, i.e., adjusting the support samples and adding pseudo support samples. Although many transductive meta-learning approaches have been proposed \cite{nichol2018first,ye2020few,yang2020dpgn,hu2020empirical}, their models are not universal and cannot be directly applied to the base learners like LSSVM.

Our first transductive module is basing on the attention mechanism \cite{vaswani2017attention}, denoted as \emph{IAM} (Inverse Attention Module). The motivation is that if we know the classification method (e.g., SVM or LSSVM) and the samples to classify, and are allowed to adjust the support (training) samples, then we can move the support samples according to the characteristics of the classifier and the query samples to get better classifier which is completely determined by the support samples. And the way of moving needs to be meta-learned and here we resort to the attention mechanism. 

Our second transductive module is used during meta-testing, denoted as \emph{PSM} (Pseudo Support Module) which calculates the prototypes of the query set with pseudo labels and uses them as new support samples. As shown in \cite{snell2017prototypical}, the class prototype is a good representation of a class. Actually, we can iterate this process multiple times and continuously increase support samples.

We denote our model as FSLSTM, the abbreviation for \emph{Few-Shot learning with LSsvm base learner and Transductive Modules}. Specifically, our contributions can be summarized as follows.
\begin{itemize}
    \item We introduce multi-class least squares support vector machine as our base learner, which can achieve better generation than existing ones with less computational overhead.
    \item We then propose two transductive modules which significantly improve the few-shot classification accuracy, especially for the difficult 1-shot setting.
    \item Experiments show our model, FSLSTM, can achieve state-of-the-art performance on \emph{miniImageNet} and \emph{CIFAR-FS}.
\end{itemize}

\section{Related Work}

Meta-learning approaches for few-shot learning aim to learn some inductive bias that generation across a distribution of tasks \cite{vilalta2002perspective} and can be broadly categorized into three groups: 1) \emph{black-box adaptation approaches} \cite{santoro2016meta,mishra2017simple} train the neural network to generate optimal model parameters; 2) \emph{optimization-based approaches} \cite{andrychowicz2016learning,finn2017model,lee2019meta,hu2020empirical} learn how to rapidly adapt a model to a given few-shot recognition task via a small number gradient descent iterations or teach the deep network to use standard machine learning tools (e.g., ridge regression, SVM) as its internal optimization; 3) \emph{metric-learning based approaches} \cite{vinyals2016matching,snell2017prototypical,DBLP:conf/aaai/ChenZWC20} learn a distance metric between a query sample and a set of support samples of a few shot task.

Many effective base learners have been proposed. Snell, Swersky, and Zemel (\citeyear{snell2017prototypical}) proposed a simple but powerful nearest-neighbor classifier which represents each class by the meaning embedding of the samples and classify query samples based on the distance to the nearest class mean. Bertinetto et al. (\citeyear{bertinetto2018meta}) used differentiable closed-form solvers (ridge regression and logistic regression) as base learner. Lee et al. (\citeyear{lee2019meta}) used discriminatively trained linear classifier (SVM and ridge regression) as the base learner. In this work, we use multi-class least squares support vector machine as the base learner which further improves accuracy and reduces computational overhead.

In order to utilize the information from the query samples, some transdutive meta-learning approaches have been proposed. Liu et al. (\citeyear{liu2018learning}) reuse the label propagation algorithm \cite{zhu2003semi} for transductive inference within each task and boosting the performance. Qiao et al. (\citeyear{qiao2019transductive}) explored the pairwise constraints and regularization prior within each task with the setup of transduction to tailor an episodic wise metric for each task. Hu et al. (\citeyear{hu2020empirical}) proposed a gradient-based method which uses the query samples to calculate the synthetic gradients to perform internal gradient updates. Yang et al. (\citeyear{yang2020dpgn}) convey both the distribution-level and instance-level relations in each few-shot learning task using dual complete graph network. Compared with these works, our transductive modules are much more simple and generic. The attention mechanism we use in the Inverse Attention Module is similar to \cite{vaswani2017attention}, but the Key and the Query are different. It can be seen as modifying the representation of support samples with query samples as context.

\section{Proposed Model}

In this section, we first derive the meta-learning framework for few-shot learning following prior work \cite{vinyals2016matching}, and then introduce the basic components of our model FSLSTM: LSSVM base learner and transductive modules, as shown in Figure \ref{model}.

\subsection{Problem Formulation}
In a few-shot classification task, we are given some training data which contains $N$ distinct, unseen classes with $K$ samples each. For a test sample, we need to classify it correctly into one of the $N$ classes. In most prior works, each task is organized as an \emph{episode} which consists of a support set $\mathcal{S}$ and a query set $\mathcal{Q}$, and a $N$-way $K$-shot task can be defined as a tuple $\{\mathcal{S},\mathcal{Q}\}$, where
\begin{align*}
  \mathcal{S}= & \{s^{(c)}\}_{c=1}^N,\quad |s^{(c)}|=K \\
  \mathcal{Q}= & \{q^{(c)}\}_{c=1}^N,\quad |q^{(c)}|=Q
\end{align*}
where $c$ is the class index and $Q$ is the number of query samples in each class. A 3-way 2-shot task is shown in Figure \ref{model}. Before classification, a backbone $f_{\theta}(\cdot)$, a CNN or ResNet \cite{he2016deep}, is needed to map the original inputs to feature representations and is expected to extract similar representations for the (support or query) samples in the same class. Then a base learner $\mathcal{A}$ output the optimal classifier $\psi$ of the task basing on the support set $\mathcal{S}$, i.e., $\psi=\mathcal{A}(\mathcal{S};\theta)$. In particular, for transductive setting, the base learner $\mathcal{A}$ also takes the query samples as input.

Meta-learning approaches aim to learn some inductive bias that generation across a distribution of tasks, so they can be considered to learn over a collection of tasks, i.e., $\mathcal{T}_{train}=\{(\mathcal{S}_i,\mathcal{Q}_i)\}_{i=1}^I$, called \emph{meta-training set}. The model is learned by minimizing generalization error across tasks given a base learner $\mathcal{A}$ and the learning objective is:
\begin{equation}\label{metatrain}
  \min_{\theta}\mathbb{E}_{\mathcal{T}_{train}}\left[\mathcal{L}^{meta}(\mathcal{Q},\psi)\right]+\mathcal{R}(\theta),\psi=\mathcal{A}(\mathcal{S};\theta)
\end{equation}
where $\mathcal{L}^{meta}$ is a loss function, e.g., cross entropy loss, and $\mathcal{R}(\theta)$ is the regularization term.

After the model is learned, its generalization is evaluated on a set of held-out tasks, called \emph{meta-testing set}, $\mathcal{T}_{test}=\{(\mathcal{S}_j,\mathcal{Q}_j)\}_{j=1}^J$ and computed as
\begin{equation}\label{metatest}
  \mathbb{E}_{\mathcal{T}_{test}}\left[\mathcal{L}^{meta}(\mathcal{Q},\psi)\right],\quad \psi=\mathcal{A}(\mathcal{S};\theta)
\end{equation}
Following prior work \cite{ravi2016optimization,finn2017model}, the stages corresponding to Equation (\ref{metatrain}) and (\ref{metatest}) are called \emph{meta-training} and \emph{meta-testing} respectively. In addition, during meta-training, a held-out \emph{meta-validation set} $\mathcal{T}_{val}$ is kept to choose the best model parameters. The categories in $\mathcal{T}_{train}$, $\mathcal{T}_{test}$ and $\mathcal{T}_{val}$ are different to each other, which makes sure the tasks in $\mathcal{T}_{test}$ and $\mathcal{T}_{val}$ are unseen for the learned model.

\begin{table*}[h]
\centering
\begin{tabular}{cccccc}
\toprule
\multicolumn{1}{c}{\multirow{2}{*}{\textbf{Model}}}&\multirow{2}{*}{\textbf{Backbone}}&\multicolumn{2}{c}{\textbf{miniImageNet}}&\multicolumn{2}{c}{\textbf{CIFAR-FS}}\\ \cmidrule(lr){3-4} \cmidrule(lr){5-6}
\multicolumn{1}{c}{} &  & \multicolumn{1}{c}{\textbf{5way-1shot}} & \multicolumn{1}{c}{\textbf{5way-5shot}} & \multicolumn{1}{c}{\textbf{5way-1shot}} & \multicolumn{1}{l}{\textbf{5way-5shot}} \\
\midrule
Reptile+BN \cite{nichol2018first}   &  Conv4    & 49.97 $\pm$ 0.0  & 65.99 $\pm$ 0.0 & -               & -               \\
Relation Net \cite{sung2018learning}&  Conv4    & 50.40 $\pm$ 0.8  & 65.30 $\pm$ 0.7 & 55.00 $\pm$ 1.0 & 69.30 $\pm$ 0.8 \\
TPN \cite{liu2018learning}          &  Conv4    & 55.51 $\pm$ 0.0  & 69.86 $\pm$ 0.0 & -               & -               \\
TEAM \cite{qiao2019transductive}    &  Conv4    & 56.57 $\pm$ 0.0  & 72.04 $\pm$ 0.0 & -               & -               \\
FEAT \cite{ye2020few}               &  Conv4    & 57.04 $\pm$ 0.2  & 72.89 $\pm$ 0.2 & -               & -               \\
FEAT$^{\ddagger}$ \cite{ye2020few}  &  Conv4    & 58.98 $\pm$ 0.2  & 74.72 $\pm$ 0.2 & -               & -               \\
SIB \cite{hu2020empirical}          &  Conv4    & 58.00 $\pm$ 0.6  & 70.70 $\pm$ 0.4 & 68.70 $\pm$ 0.6 & 77.10 $\pm$ 0.4 \\
SIB$^{\ddagger}$ \cite{hu2020empirical}&  Conv4    & \textbf{65.04 $\pm$ 0.8}  & \textbf{77.20 $\pm$ 0.5} & 74.10 $\pm$ 0.9 & 82.78 $\pm$ 0.5 \\
\hline
\textbf{LSSVM}$^{\ddagger}$     &  Conv4    & 58.13 $\pm$ 0.6  & 75.09 $\pm$ 0.4 & 69.80 $\pm$ 0.7 & 81.43 $\pm$ 0.5 \\
\textbf{LSSVM+BN}$^{\ddagger}$  &  Conv4    & 57.99 $\pm$ 0.6  & 75.32 $\pm$ 0.4 & 69.99 $\pm$ 0.7 & 82.48 $\pm$ 0.5 \\
\textbf{LSSVM+IAM}$^{\ddagger}$ &  Conv4    & 59.29 $\pm$ 0.6  & 76.70 $\pm$ 0.4 & 71.99 $\pm$ 0.7 & 85.36 $\pm$ 0.5 \\
\textbf{LSSVM+IAM+PSM (FSLSTM)}$^{\ddagger}$&  Conv4    & 62.98 $\pm$ 0.6  & \textbf{77.72 $\pm$ 0.4} & \textbf{78.18 $\pm$ 0.7} & \textbf{86.41 $\pm$ 0.5} \\
\hline
FEAT \cite{ye2020few}               & ResNet12  & 66.78 $\pm$ 0.2  & 82.05 $\pm$ 0.2 & -               & -               \\
FEAT$^{\ddagger}$ \cite{ye2020few}  & ResNet12  & 69.96 $\pm$ 0.2  & 84.32 $\pm$ 0.2 & -               & -                  \\
SIB \cite{hu2020empirical}          & ResNet12  & 70.40 $\pm$ 0.8  & 80.16 $\pm$ 0.5 & 77.04 $\pm$ 0.8 & 84.25 $\pm$ 0.6  \\
SIB$^{\ddagger}$ \cite{hu2020empirical}& ResNet12  & \textbf{74.80 $\pm$ 0.8}  & 83.65 $\pm$ 0.5 & 82.17 $\pm$ 0.7 & 88.05 $\pm$ 0.5   \\
DPGN \cite{yang2020dpgn}            & ResNet12  & 67.77 $\pm$ 0.3  & 84.60 $\pm$ 0.4 & 77.90 $\pm$ 0.5 & 90.20 $\pm$ 0.4 \\
DPGN$^{\ddagger}$ \cite{yang2020dpgn}& ResNet12 & 69.54 $\pm$ 0.5  & 85.72 $\pm$ 0.4 & 80.14 $\pm$ 0.5 & 91.83 $\pm$ 0.3  \\
\hline
\textbf{LSSVM}$^{\ddagger}$     & ResNet12  & 68.46 $\pm$ 0.6  & 85.14 $\pm$ 0.4 & 78.60 $\pm$ 0.6 & 91.17 $\pm$ 0.4    \\
\textbf{LSSVM+BN}$^{\ddagger}$  & ResNet12  & 69.48 $\pm$ 0.6  & 85.46 $\pm$ 0.4 & 81.09 $\pm$ 0.6 & 91.40 $\pm$ 0.4  \\
\textbf{LSSVM+IAM}$^{\ddagger}$ & ResNet12  & 70.96 $\pm$ 0.6  & 85.99 $\pm$ 0.4 & 81.66 $\pm$ 0.6 & 92.01 $\pm$ 0.4  \\
\textbf{LSSVM+IAM+PSM (FSLSTM)}$^{\ddagger}$& ResNet12  & \textbf{75.54 $\pm$ 0.6}  & \textbf{86.75 $\pm$ 0.4} & \textbf{86.88 $\pm$ 0.6}& \textbf{92.82 $\pm$ 0.4} \\
\bottomrule
\end{tabular}
\caption{Comparison to previous transductive meta-learning approaches on \emph{miniImageNet} and \emph{CIFAR-FS}. 'LSSVM' represents using multi-class LSSVM as the base learner. 'BN' represents information is shared among samples using batch normalization. '${\ddagger}$' stands for using the weights pretrained on the Places365-standard dataset as initialization. The results of the baselines are reported by \cite{liu2018learning,qiao2019transductive,ye2020few,hu2020empirical} and our reimplementation.}
\label{SOTA}
\end{table*}
\subsection{LSSVM Base Learner}\label{LS_SVM}
The choice of base learner $\mathcal{A}$ is crucial to few-shot classification. Many choices of base learners have been proposed \cite{snell2017prototypical,bertinetto2018meta,gidaris2018dynamic,lee2019meta}, among them MetaOptNet \cite{lee2019meta} achieve impressive performance
using discriminatively trained linear classifier (e.g., SVM). However, it needs to solve the quadratic programming problem with iterative algorithm which brings an increase in computational overhead. In addition, due to the limitation of the number of iterations and the quadratic programming solver, the obtained classifier can only be approximately optimal. Instead, we use multi-class least square support vector machine as our base learner which only need to solve the system of linear equations obtained from the Karush-Kuhn-Tucker (KKT) condition and can get the optimal solution.

There are three basic coding approaches for multi-class problems using binary classifiers: one-vs-all, one-vs-one and error correcting output codes (ECOC) \cite{garcia2011empirical}. Given train set $\{(x_i,y_i)\}_{i=1}^n$ with $C$ distinct categories, coding each category with a vector of length $L$ with each element chosen from $\{0,\pm1\}$, we can get a coding matrix $M\in\mathbb{R}^{C\times L}$ and $L$ train sets $\{(x_i,y_i^l)\}_{i=1}^n,l=1,\cdots,L$ for $L$ binary classifiers respectively. The quadratic programming formulation is 
\begin{align}\label{LSSVM}
  \min_{w,b}\quad & \sum_{l=1}^{L}\left[\frac{1}{2}(w_l^Tw_l+b_l^2)+\frac{\gamma}{2}\sum_{i=1}^{n}e_i^{l2}\right] \\
  s.t.\quad & y_i^l\left[w_l^T\varphi(x_i)+b_l\right]=1-e_i^l,\forall i,l
\end{align}
where $\gamma$ is the regularization parameter and $\varphi(\cdot)$ is a mapping function. Using the Karush-Kuhn-Tucker (KKT) condition, we only need to solve the system of linear equations:
\begin{equation}\label{solve1}
  \begin{bmatrix}
    -2I & Y^T \\
    Y & \Omega
  \end{bmatrix}\begin{bmatrix}
                 \textbf{b} \\
                 \bm{\alpha}
               \end{bmatrix}=\begin{bmatrix}
                               \textbf{0} \\
                               \textbf{1}
                             \end{bmatrix}
\end{equation}
where $I$ is the identity matrix, $\bm{\alpha}=[\alpha_1^T,\cdots,\alpha_L^T]^T$ is the dual variable and
\begin{equation}\label{solve2}
\Omega=\begin{bmatrix}
         \Omega_1 & \cdots & 0 \\
         \vdots   & \ddots & \vdots \\
                0 & \cdots & \Omega_L
       \end{bmatrix}\qquad Y=\begin{bmatrix}
                               y^1    & \cdots & 0 \\
                               \vdots & \ddots & \vdots \\
                               0      & \cdots & y^L
                             \end{bmatrix}
\end{equation}
and $y^l_i\in\{0,\pm1\},\Omega_l^{ij}=y_i^ly_j^l\Phi(x_i,x_j)+\frac{1}{\gamma}\delta_{ij},\forall i,j,l$ with $\Phi(\cdot,\cdot)$ is the kernel function and $\delta_{ij}$ is the Kronecker delta function. Given a test sample $x$, its classification results in the $L$ classifiers constitute its class code and take the class with the smallest code Hamming distance as the prediction class, i.e.,
\begin{align}\label{predict}
    y=     & \arg\min_r\sum_{l=1}^{L}\frac{1-sgn(M_{rl}\cdot sgn(c_l(x)))}{2} \\
   \approx & \arg\max_r \sum_{l=1}^{L}M_{rl}\cdot c_l(x),r=1,\cdots,C
\end{align}
where $c_l(x)=\sum_{i=1}^{n}\alpha_l^iy^l_i\Phi(x_i,x)+b_l$ and $sgn(\cdot)$ is the sign function.

Note that there are only $L(n+1)$ variables in Equation (\ref{solve1}) which is small in the few-shot task ($n=NK$), and there is no need to iterate multiple steps. Using this base learner, our system is end-to-end trainable which means we need to calculate $\frac{\partial \alpha}{\partial x}$ and $\frac{\partial b}{\partial x}$, where $x$ corresponds to the output of the backbone. Using implicit function theorem \cite{dontchev2009implicit,krantz2012implicit} on Equation (\ref{solve1}), we can solve it easily.

\begin{table*}[h]
\center
\begin{tabular}{ccccccccc}
\hline
\multirow{3}{*}{\textbf{Base Learner}} & \multicolumn{4}{c}{\textbf{miniImageNet}} & \multicolumn{4}{c}{\textbf{CIFAR-FS}} \\ \cmidrule(lr){2-5} \cmidrule(lr){6-9} 
& \multicolumn{2}{c}{\textbf{5way-1shot}}   & \multicolumn{2}{c}{\textbf{5way-5shot}} & \multicolumn{2}{c}{\textbf{5way-1shot}} & \multicolumn{2}{c}{\textbf{5way-5shot}} \\
& \multicolumn{1}{c}{Acc} & Time  & \multicolumn{1}{c}{Acc} & \multicolumn{1}{c}{Time} & Acc  & \multicolumn{1}{c}{Time} & \multicolumn{1}{c}{Acc} & Time  \\
\hline
NN      & 59.25 $\pm$ 0.6          & 286 & 75.60 $\pm$ 0.5          & 358  & \textbf{72.20 $\pm$ 0.7} & 115 & 83.50 $\pm$ 0.5          & 146 \\
RR      & 61.41 $\pm$ 0.6          & 524 & 77.88 $\pm$ 0.5          & 600  & \textbf{72.60 $\pm$ 0.7} & 335 & 84.30 $\pm$ 0.5          & 371  \\
SVM     & \textbf{62.64 $\pm$ 0.6} & 732 & \textbf{78.63 $\pm$ 0.5} & 1062 & 72.00 $\pm$ 0.7          & 629 & 84.20 $\pm$ 0.5          & 884 \\
LSSVM   & \textbf{63.22 $\pm$ 0.6} & 303 & \textbf{79.02 $\pm$ 0.5} & 387  & \textbf{73.40 $\pm$ 0.7} & 132 & \textbf{85.49 $\pm$ 0.5} & 155  \\ 
\hline
\end{tabular}
\caption{Average few-shot classification accuracy($\%$) with $95\%$ confidence intervals and time(s) required to solve 10,000 randomly sampled tasks from \emph{miniImageNet} and \emph{CIFAR-FS} on a single NVIDIA RTX 2080Ti GPU. Here 'NN' stands for nearest-neighbor classifier \cite{snell2017prototypical} and 'RR' stands for ridge regression \cite{lee2019meta}. The accuracy  of the baselines is reported from \cite{lee2019meta}.}
\label{acc_speed}
\end{table*}
\subsection{Transductive Modules}\label{TML}
Next, we introduce how to use query samples to modify the support set so as to obtain the better classifier parameters.

\subsubsection{Inverse Attention Module}
As we can see, Equation (\ref{solve1}) has a unique solution, so given the support set, our base learner outputs the unique optimal classifier and the decision boundary changes with the support set. Intuitively, the support samples can be adjusted using the query samples to obtain the better classifier parameters.

Specifically, we use the attention mechanism \cite{vaswani2017attention} to introduce knowledge from the query samples. Let $f_{\theta}(\mathcal{S})$ and $f_{\theta}(\mathcal{Q})$ be the feature vectors of the support set $\mathcal{S}$ and the query set $\mathcal{Q}$ respectively, define (query $\mathbf{Q}$, key $\mathbf{K}$, value $\mathbf{V}$) as
\begin{align}\label{QKV}
  \mathbf{Q}= & g_{\phi}^q(f_{\theta}(\mathcal{S}))\in\mathbb{R}^{NK\times d_k}\\
  \mathbf{K}= & g_{\phi}^k(f_{\theta}(\mathcal{Q}))\in\mathbb{R}^{NQ\times d_k}\\
  \mathbf{V}= & g_{\phi}^v(f_{\theta}(\mathcal{Q}))\in\mathbb{R}^{NQ\times d_v}
\end{align}
where $g_{\phi}^q(\cdot)$, $g_{\phi}^k(\cdot)$ and $g_{\phi}^v(\cdot)$ are three different mapping functions. Once we have $(\mathbf{Q},\mathbf{K},\mathbf{V})$, we use them to compute the Scaled Dot-Product Attention:
\begin{equation}\label{attention}
  A(\mathbf{Q},\mathbf{K},\mathbf{V})=softmax\left(\frac{\mathbf{Q}\mathbf{K}^T}{\sqrt{d_k}}\right)\mathbf{V}
\end{equation}
Using the labels from the support set, we can know the class corresponding to each row of the matrix $A$. To increase the intra-class similarity, we replace each row of $A$ with the prototype of its corresponding class which is calculated from the matrix $A$. Finally, we can calculate the adjusted support set as
\begin{equation}\label{support}
  \mathcal{S}:=LN(\mathcal{S}+Dropout(h_{\phi}(A(\mathbf{Q},\mathbf{K},\mathbf{V}))))
\end{equation}
where $h_{\phi}$ is another mapping function and $LN$ represents layer normalization \cite{ba2016layer}.

Generally, the mapping functions $g_{\phi}^q(\cdot)$, $g_{\phi}^k(\cdot)$ and $g_{\phi}^v(\cdot)$ are linear functions defined by a weight matrix $W$. To limit model complexity while increasing model capability, we use the bottleneck with two fully connected layers, i.e., a dimension-reduction layer with parameters $W_1$ and reduction ratio $r$, a ReLU and a dimension-increasing layer with parameters $W_2$, as the mapping functions. This module is called Inverse Attention Module since it uses the support set as the Query and uses the query set as the Key and the Value. It uses the weighted combination of the Value as the offset of the support samples and the weight is calculated basing on the relationship between the Key and the Query.

\subsubsection{Pseudo Support Module}
In a few-shot classification task, the support samples and the query samples satisfy the \emph{clustering assumption}. So when the original classifier has been able to classify samples well, the prototypes of the query set with pseudo labels can be used as the effective support samples, and they have a closer relationship with the query samples. We can use them to enhance classification.

Let the query set be $\mathcal{Q}=\{x_i\}_{i=1}^{NQ}$, using the optimal classifier from our base learner, we can get new query set with pseudo labels $\widetilde{\mathcal{Q}}=\{(x_i,\widetilde{y_i})\}_{i=1}^{NQ}$. Its prototypes can be computes as
\begin{equation}\label{proto}
  p_k=\frac{1}{|Q_k|}\sum_{x_i\in Q_k}f_{\theta}(x_i),k=1,\cdots,N
\end{equation}
where $Q_k=\{x|(x,\widetilde{y})\in \widetilde{\mathcal{Q}},\widetilde{y}=k\}$. Note that this operation can be iterated multiple times and is only used during meta-testing to avoid increasing computation overhead during meta-training.

\section{Experiments}
In this section, we first briefly describe our experimental setting. Next, we compare our model, FSLSTM, with the existing state-of-the-art transductive meta-learning approaches. Then we compare our LSSVM base learner with the existing ones. After that, we show the robustness of our Pseudo Support Module. Finally, we analyze the proposed two transductive modules in detail.

\begin{table*}[h]
\center
\begin{tabular}{cccccc}
\toprule
\multicolumn{1}{c}{\multirow{2}{*}{\textbf{Base Learner}}} & \multicolumn{1}{c}{\multirow{2}{*}{\textbf{Backbone}}} & \multicolumn{2}{c}{\textbf{miniImageNet}} & \multicolumn{2}{c}{\textbf{CIFAR-FS}} \\
\cmidrule(lr){3-4} \cmidrule(lr){5-6}
\multicolumn{1}{c}{} & & \multicolumn{1}{c}{\textbf{5way-1shot}} & \multicolumn{1}{c}{\textbf{5way-5shot}} & \multicolumn{1}{c}{\textbf{5way-1shot}} & \multicolumn{1}{c}{\textbf{5way-5shot}} \\
\midrule
NN        & ResNet12 & 59.25 $\pm$ 0.6          & 75.60 $\pm$ 0.5             & 72.20 $\pm$ 0.7          & \textbf{83.50 $\pm$ 0.5} \\
NN+CAN    & ResNet12 & 62.63 $\pm$ 0.7          & \textbf{76.99 $\pm$ 0.5}    & \textbf{77.72 $\pm$ 0.8} & \textbf{84.07 $\pm$ 0.5} \\
NN+PSM    & ResNet12 & \textbf{65.94 $\pm$ 0.7} & \textbf{77.03 $\pm$ 0.5}    & \textbf{78.91 $\pm$ 0.8} & \textbf{84.34 $\pm$ 0.5} \\
\hline
RR        & ResNet12 & 61.41 $\pm$ 0.6          & 77.88 $\pm$ 0.5             & 72.60 $\pm$ 0.7          & 84.30 $\pm$ 0.5 \\
RR+CAN    & ResNet12 & 59.77 $\pm$ 0.7          & 74.86 $\pm$ 0.5             & 72.33 $\pm$ 0.8          & 83.11 $\pm$ 0.6 \\
RR+PSM    & ResNet12 & \textbf{66.51 $\pm$ 0.7} & \textbf{78.96 $\pm$ 0.5}    & \textbf{78.68 $\pm$ 0.8} & \textbf{85.47 $\pm$ 0.5} \\
\hline
SVM       & ResNet12 & 62.64 $\pm$ 0.6          & \textbf{78.63 $\pm$ 0.5}    & 72.00 $\pm$ 0.7          & 84.20 $\pm$ 0.5 \\
SVM+CAN   & ResNet12 & 62.14 $\pm$ 0.7          & 78.33 $\pm$ 0.5             & 71.37 $\pm$ 0.8          & 83.76 $\pm$ 0.5 \\
SVM+PSM   & ResNet12 & \textbf{66.14 $\pm$ 0.7} & \textbf{79.34 $\pm$ 0.5}    & \textbf{77.79 $\pm$ 0.8} & \textbf{85.22 $\pm$ 0.5} \\
\hline
LSSVM     & ResNet12 & 63.22 $\pm$ 0.6          & \textbf{79.02 $\pm$ 0.5}    & 73.40 $\pm$ 0.7          & \textbf{85.49 $\pm$ 0.5} \\
LSSVM+CAN & ResNet12 & 62.43 $\pm$ 0.7          & 77.66 $\pm$ 0.5             & 72.72 $\pm$ 0.7          & 84.96 $\pm$ 0.5 \\
LSSVM+PSM & ResNet12 & \textbf{66.26 $\pm$ 0.7} & \textbf{79.46 $\pm$ 0.5}    & \textbf{78.29 $\pm$ 0.7} & \textbf{86.29 $\pm$ 0.5} \\
\bottomrule
\end{tabular}
\caption{Average few-shot classification accuracy with $95\%$ confidence intervals on \emph{miniImageNet} and \emph{CIFAR-FS} with ResNet12 as the backbone. 'CAN' means using the transductive method in \cite{hou2019cross}.}
\label{CAN}
\end{table*}
\subsection{Experimental Setting}\label{ID}

We evaluate our model on two standard few-shot learning benchmarks: \emph{miniImageNet} \cite{vinyals2016matching} and \emph{CIFAR-FS} \cite{bertinetto2018meta}.

For fair and comprehensive comparison with previous approaches, we employ two popular networks as our backbone: 1) Conv4 \cite{kim2019edge,ye2020few,yang2020dpgn} consists of four Conv-BN-ReLU blocks and the last two blocks contain the dropout layer \cite{srivastava2014dropout}; 2) ResNet12 \cite{lee2019meta,ye2020few,yang2020dpgn} consists of four residual blocks and each residual block consists of three Conv-BN-ReLU blocks. Instead of optimizing from scratch, we apply an additional pretraining strategy as in \cite{fei2020meta} which pretrains the backbone using Places365-standard dataset \cite{zhou2017places} for the standard 365-way classification. The purpose of using the pretraining strategy is to warm up the backbone and thus assist training the transductive modules. During pretraining, we crop the images to $84\times84$ and $32\times32$ for \emph{miniImageNet} and \emph{CIFAR-FS} respectively. We use SGD with Nesterov momentum of 0.9 and weight decay of 0.0005. The model is meta-trained for 60 epochs and each epoch consists of 1000 batches with each batch consisting of 8 episodes. Considering the Pseudo Support Module can be performed multiple times, we choose the number of iterations $k=10$ during meta-testing for all experiments.

We evaluate our model in 5-way 1-shot/5-shot settings and randomly sample 1000 episodes for evaluation and report the average accuracy ($\%$) as well as $95\%$ confidence interval.

Please see the supplementary material for more details.

\subsection{Comparison with State-of-the-arts}\label{TM}

We compare our model with several state-of-the-art transductive meta-learning approaches, and the result is shown in Table \ref{SOTA}. We use the same pretraining strategy as us for several baselines for fair comparison. Specifically, for FEAT \cite{ye2020few} which uses a different pretraining strategy, we directly replace its original pretraining weights with the weights obtained by our pretraining strategy. For SIB \cite{hu2020empirical} which freezes the backbone during meta-training, if we directly use the weights pretrained on Places365-standard dataset, the final result is very poor because there is no information about \emph{miniImageNet} or \emph{CIFAR-FS}, so we first use the weights pretrained on Places365-standard dataset as initialization, and then pretrain the backbone again on \emph{miniImageNet} and \emph{CIFAR-FS} using the strategy in SIB. For DPGN \cite{yang2020dpgn} which does not use any pretraining strategy, we use the weights obtained by our pretraining strategy as the initialization of the backbone and reduce its learning rate. All results are obtained using the public implementation published by the authors.

As shown in Table \ref{SOTA}, the new pretraining strategy improves the few-shot classification accuracy to different degrees, which shows the potential of transfer learning to improve few-shot learning. Furthermore, our first transductive module, IAM, can achieve the improvements ranging from $0.84\%$ to $3.93\%$ over the LSSVM base learner, and outperforms batch normalization. Our second transductive module, PSM, further improve the few-shot classification accuracy, especially for difficult 1-shot tasks with the improvement ranging from $3.69\%$ to $6.19\%$. Those results clearly validate the effectiveness of our transductive modules and our model, FSLSTM (LSSVM+IAM+PSM), achieves state-of-the-art performance on both datasets.

\subsection{Base Learner Comparison}\label{BBL}

Next, we compare the LSSVM base learner with existing ones, i.e., nearest-neighbor classifier \cite{snell2017prototypical}, ridge-regression \cite{bertinetto2018meta,lee2019meta} and SVM \cite{lee2019meta}, in terms of accuracy and inference speed. For fair comparisons, we employ them on \emph{miniImageNet} and \emph{CIFAR-FS} using the same backbone without pretraining. Specifically, we use ResNet12 as the backbone to evaluate accuracy, and use Conv4 as the backbone to evaluate inference speed. Using the small convolutional network can better reflect the difference in inference speed among different base learners. We compare the amount of time required to solve 10,000 randomly sampled tasks on a single NVIDIA RTX2080Ti GPU.

As shown in Table \ref{acc_speed}, the LSSVM base learner achieves higher accuracy than existing ones with faster inference speed. Specifically, the LSSVM base learner outperforms MetaOptNet-SVM \cite{lee2019meta} on both few-shot learning benchmarks and is $2\sim6$ times faster than it with Conv4 as the backbone. On the other hand, the LSSVM base learner is as fast as Prototypical Network \cite{snell2017prototypical} but achieves significantly better results.

\subsection{Robustness of Pseudo Support Module}
A closely related work \cite{hou2019cross} with our Pseudo Support Module augments the support set using the confidently classified query images during meta-testing, i.e., choosing the query samples with higher predicted scores as the pseudo support samples. Similar to our module, their method can also be iterated multiple times. It is worth noting that their method is specifically designed for metric-learning based approaches, i.e., matching network \cite{vinyals2016matching}, prototypical network \cite{snell2017prototypical} and relation network \cite{sung2018learning}. So it is interesting to explore the effect of this method on other base learners (e.g., ridge-regression, SVM and LSSVM), and can be directly compared with our module.

For a fair comparison, we use the transductive operation in CAN \cite{hou2019cross} and our Pseudo Support Module for various base learners, i.e., nearest-neighbor classifier \cite{snell2017prototypical}, ridge-regression \cite{lee2019meta}, SVM \cite{lee2019meta} and the LSSVM base learner. We use ResNet12 as the backbone and do not apply the pre-training initialization. We implement the transductive operation in CAN \cite{hou2019cross} for two iterations with 35 candidates for the first iteration and 70 for the second iteration, as suggested by the authors.

The results are shown in Table \ref{CAN}. From these results, we can make following observations: 1) our Pseudo Support Module achieves impressive effect on various base learners, not only the metric-learning based ones, but also the optimization-based ones; 2) the transductive method in CAN \cite{hou2019cross} significantly improves the metric-learning based approaches, but reduces the accuracy of the optimization-based approaches, and the reason may be that in the metric-learning based approaches, the predicted scores are directly related to the confidence of the candidate samples, while other optimization-based approaches have more complex classification mechanisms and the predicted scores and the confidence are no longer directly related, so the candidate samples become noise; 3) our transductive operation consistently outperforms the method in \cite{hou2019cross}, even in metric-learning based approaches, which means our method is more universal and effective, and the class prototypes are more statistically valid and robust compared with specific samples.

\begin{figure}[h]
    \centering
    \subfigure[Acc: $65.3\%\rightarrow70.6\%$]{
        \includegraphics[width=0.22\textwidth]{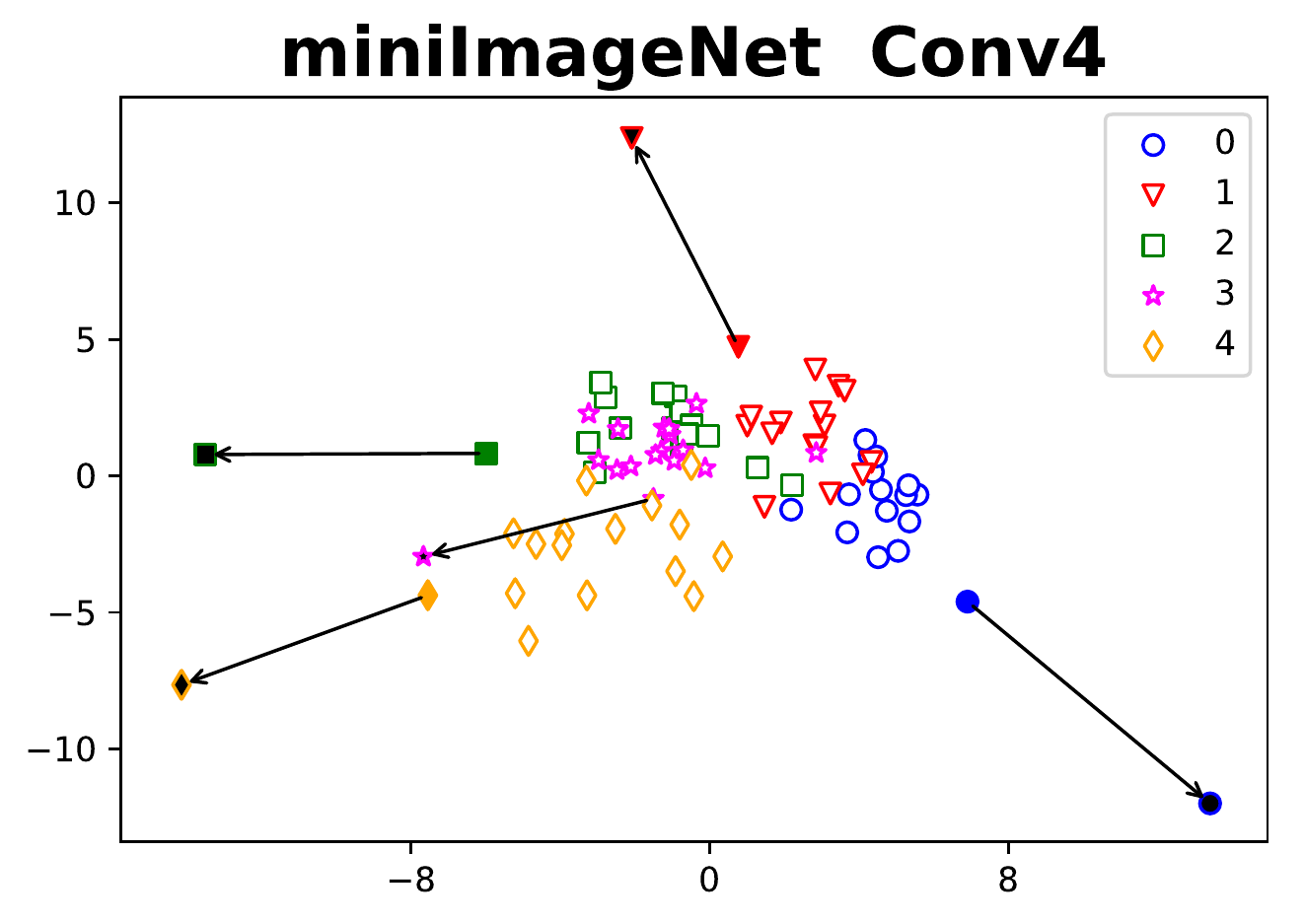}}
    \subfigure[Acc: $76.0\%\rightarrow80.0\%$]{
        \includegraphics[width=0.22\textwidth]{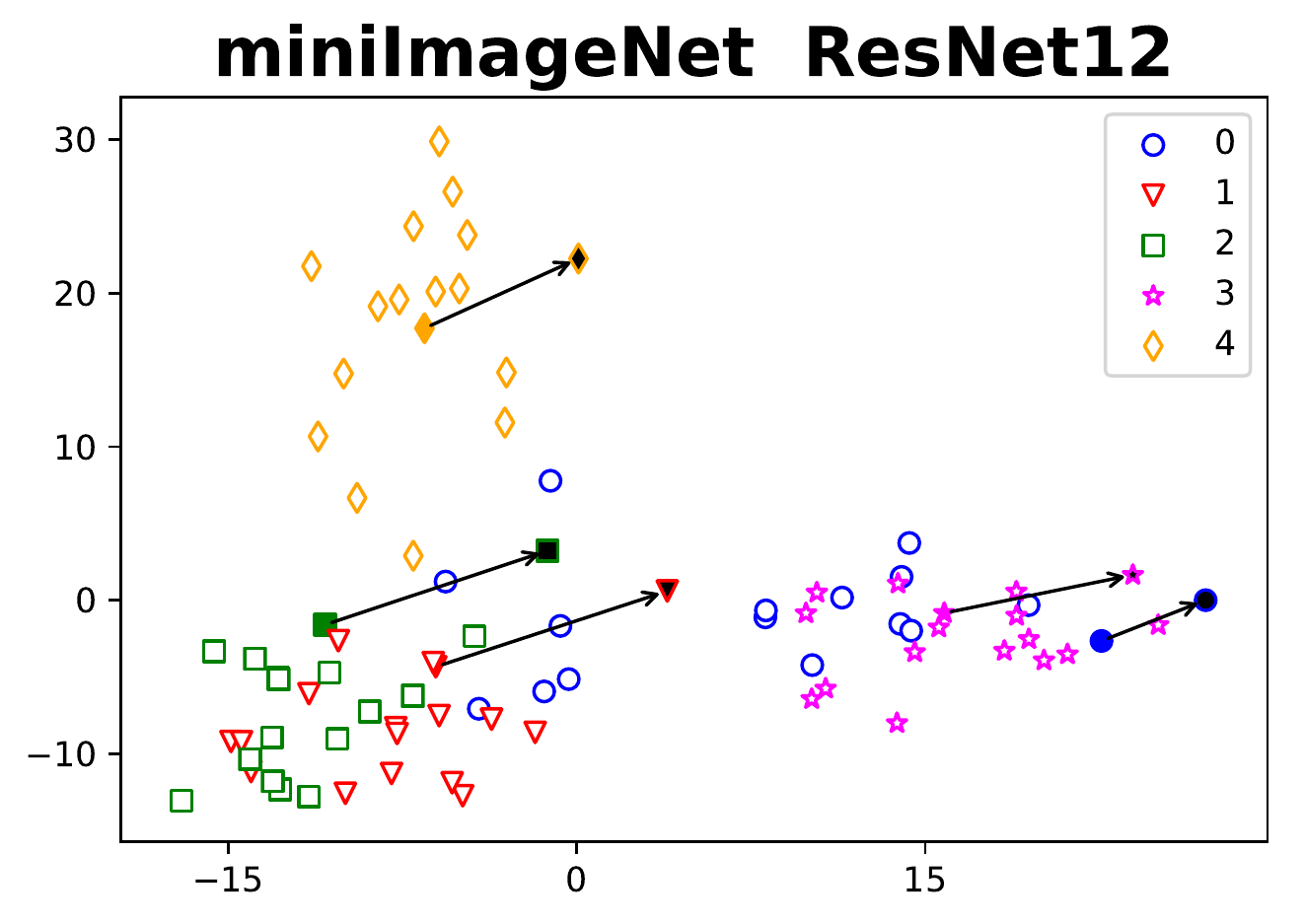}}
    \subfigure[Acc: $72.0\%\rightarrow77.3\%$]{
        \includegraphics[width=0.22\textwidth]{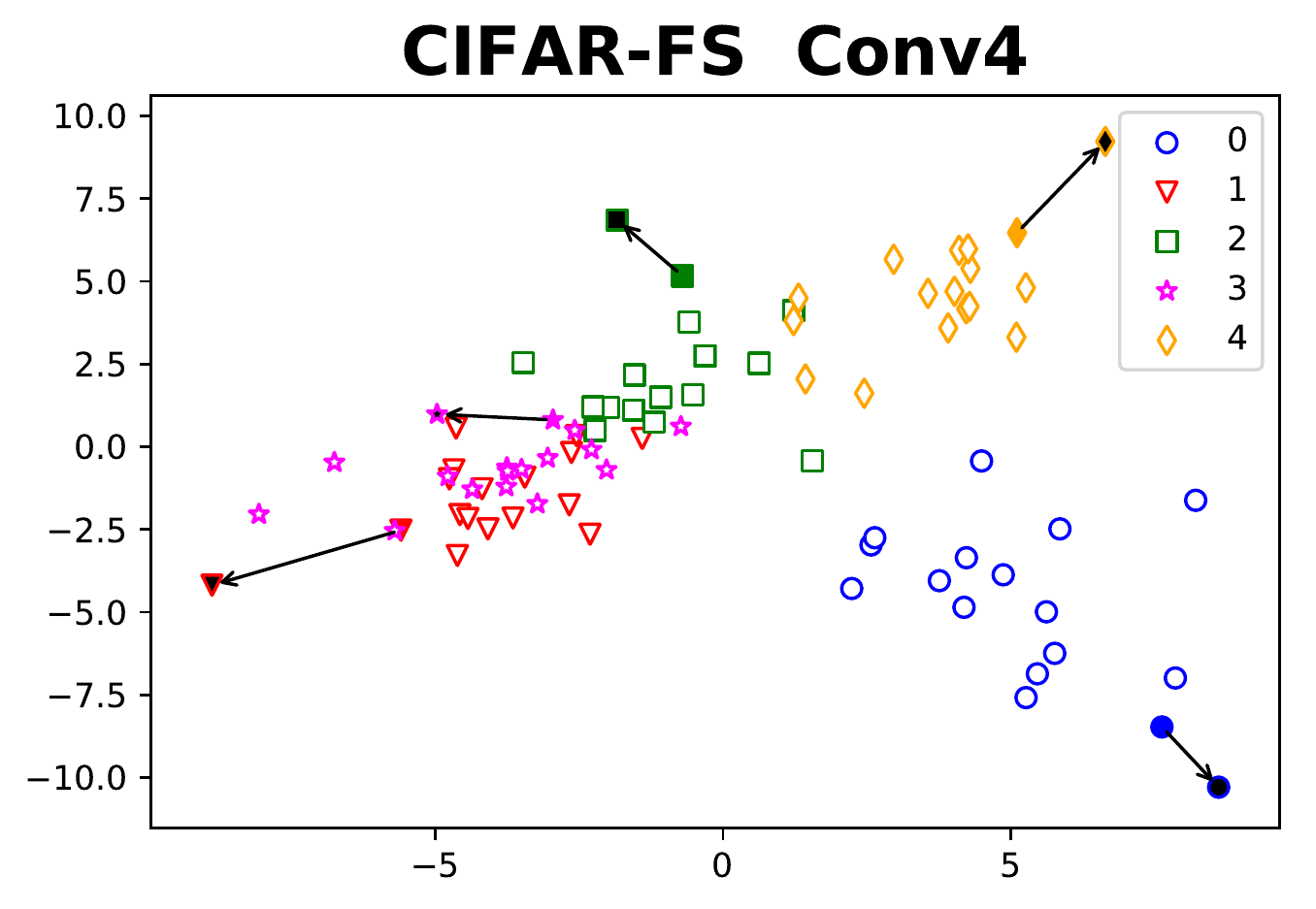}}
    \subfigure[Acc: $94.6\%\rightarrow93.3\%$]{
        \includegraphics[width=0.22\textwidth]{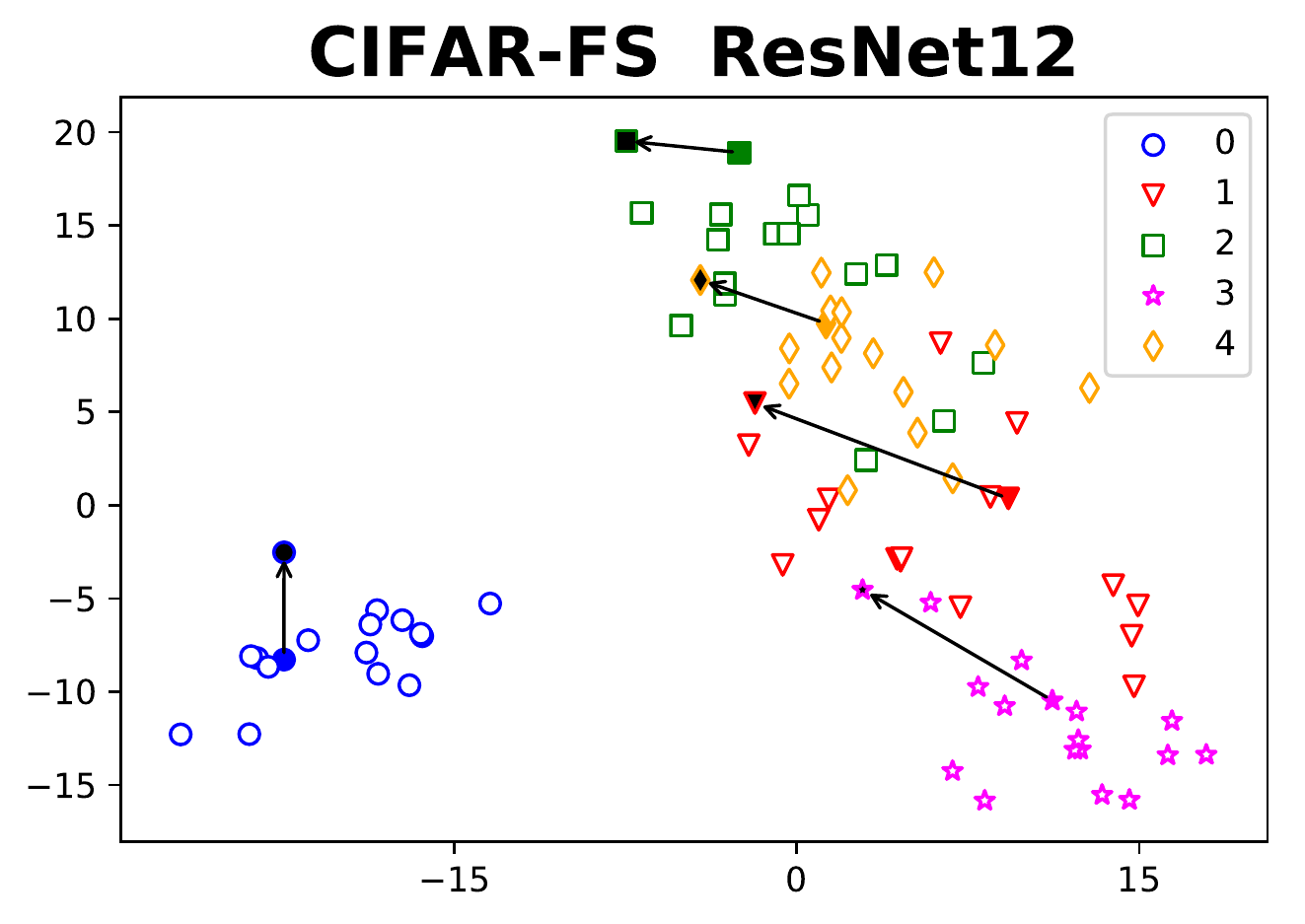}}
    \caption{Change of the support samples in 5-way 1-shot tasks under four different settings. There are 85 scatters in each figure which consist of 5 support samples, 5 adjusted support samples and 75 query samples. The arrow in each figure shows the change of the support samples before and after applying Inverse Attention Module. Values below are the classification accuracy of these four tasks before and after applying Inverse Attention Module.}
    \label{change}
\end{figure}
\begin{figure}[h]
    \centering
    \subfigure[miniImageNet  Conv4]{
        \includegraphics[width=0.22\textwidth]{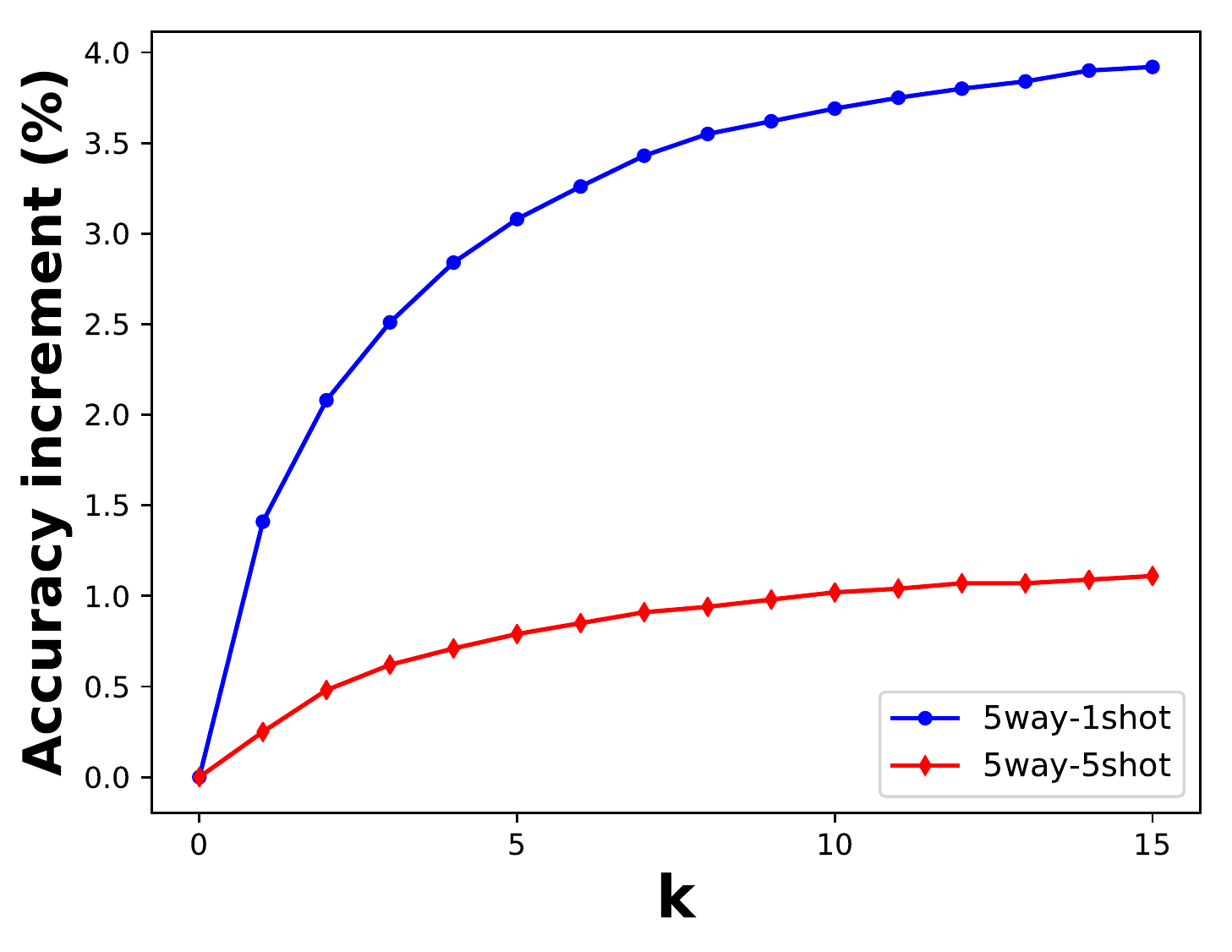}}
    \subfigure[miniImageNet  ResNet12]{
        \includegraphics[width=0.22\textwidth]{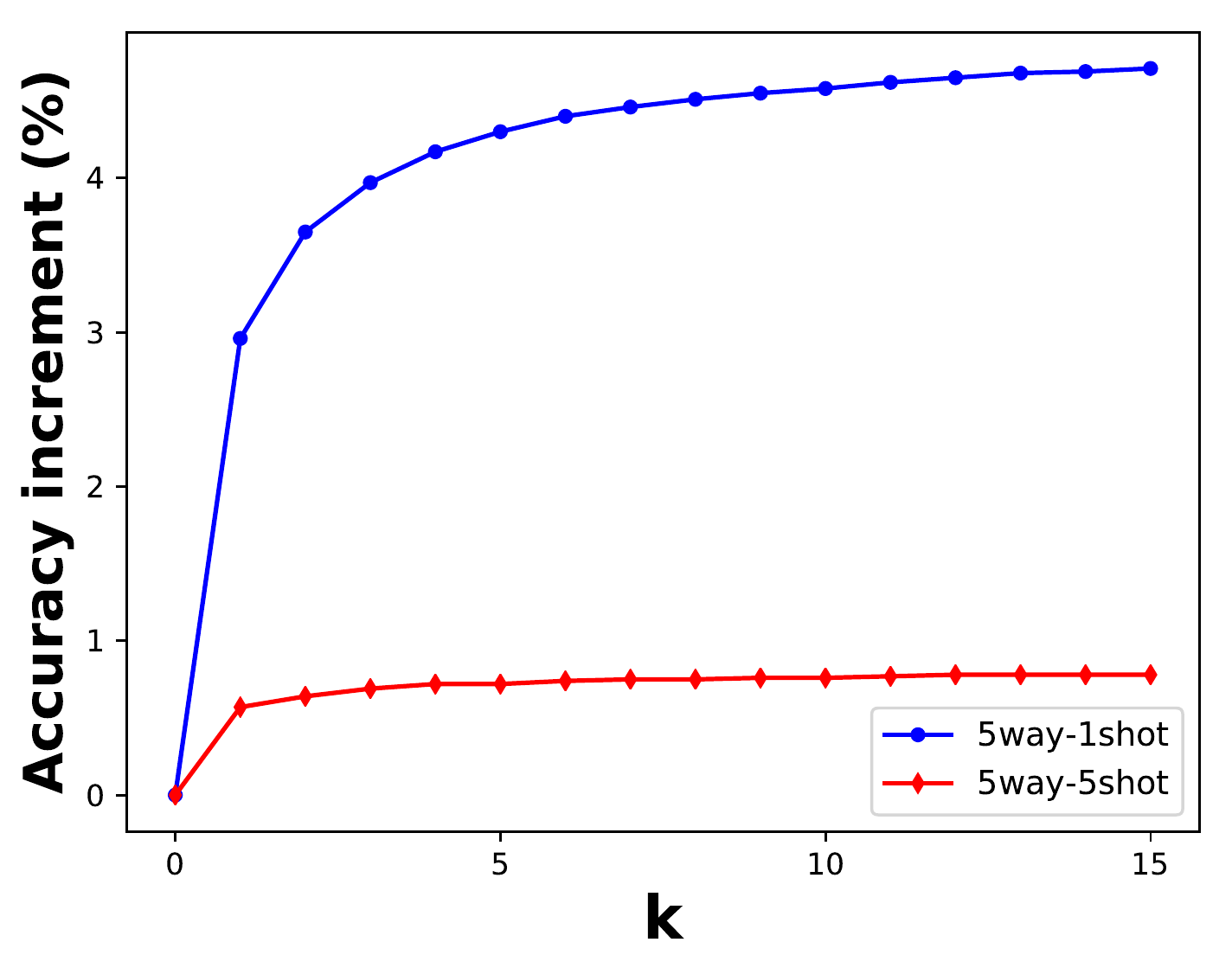}}
    \subfigure[CIFAR-FS  Conv4]{
        \includegraphics[width=0.22\textwidth]{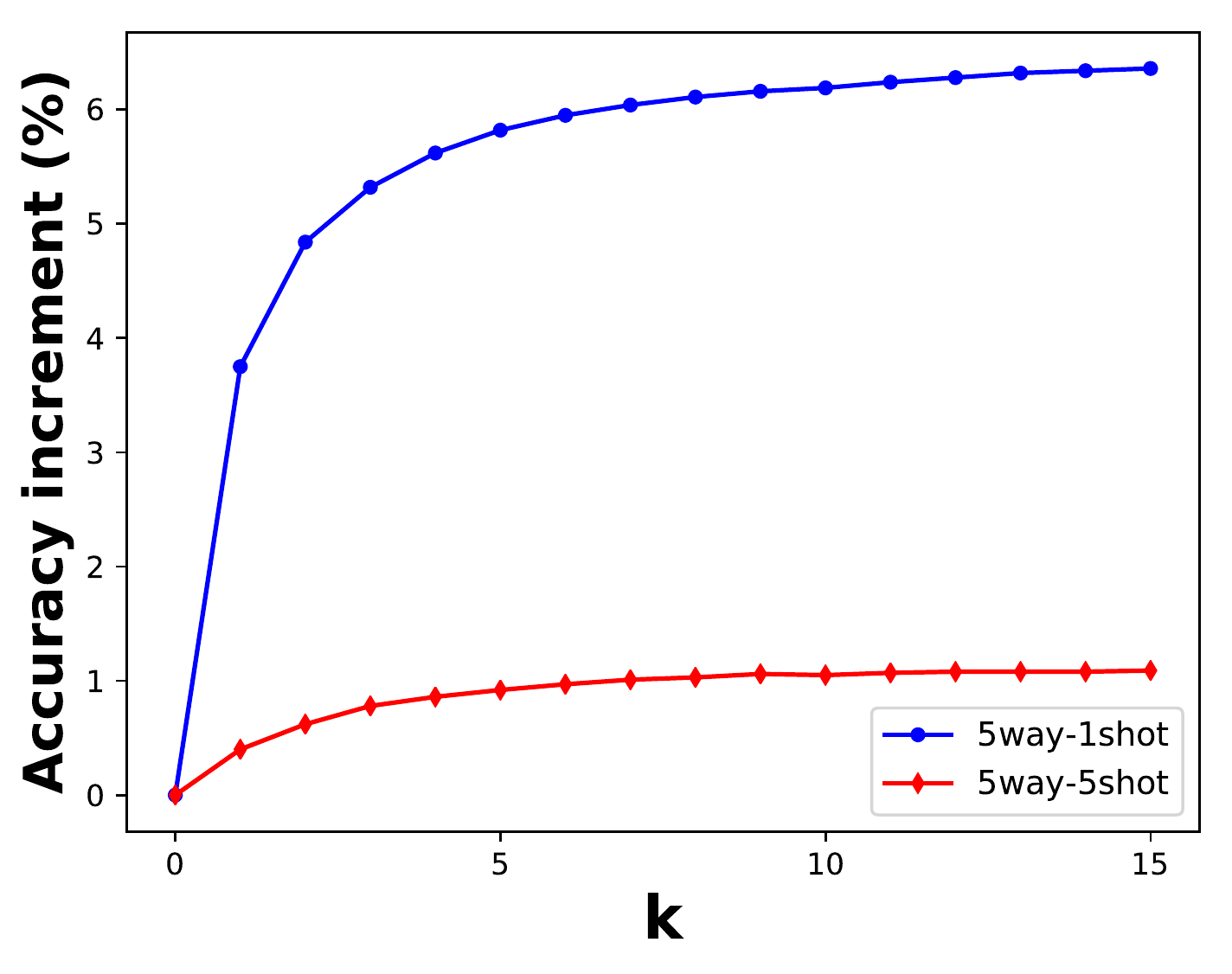}}
    \subfigure[CIFAR-FS  ResNet12]{
        \includegraphics[width=0.22\textwidth]{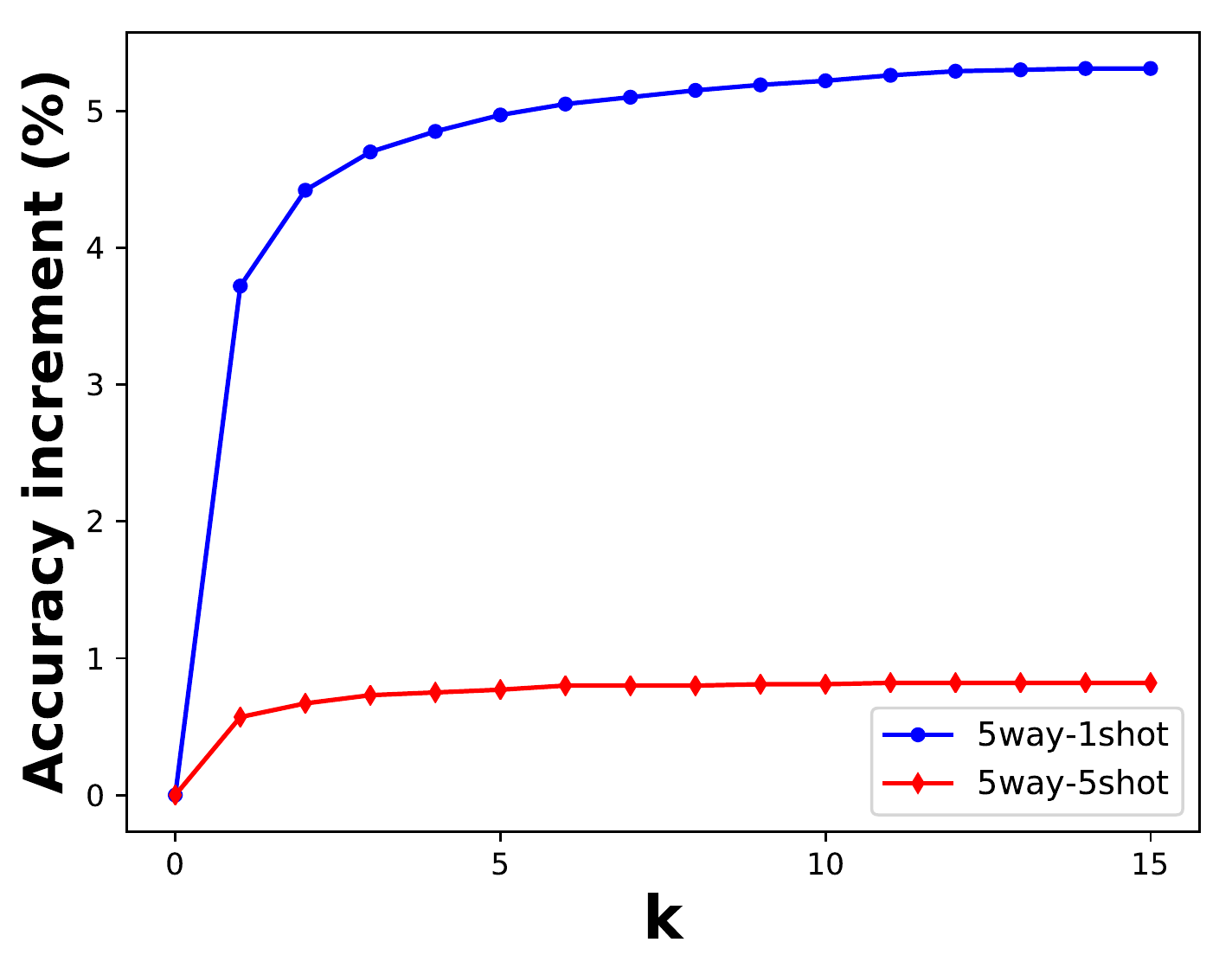}}
    \caption{Variation of the accuracy increment with the iterations number $k$ under four different settings: ResNet12 on \emph{miniImageNet}, Conv4 on \emph{miniImageNet}, ResNet12 on \emph{CIFAR-FS} and Conv4 on \emph{CIFAR-FS}.}
    \label{iteration}
\end{figure}
\subsection{Analysis on Transductive Modules}\label{AS}

Finally, we analyze our transductive modules in detail. We perform qualitative study on the Inverse Attention Module, visualizing the change of the support samples. Then we perform quantitative study on the Pseudo Support Module to explore the impact of its iteration number on accuracy.

\subsubsection{Change of the support samples}
We randomly sample four 5-way 1-shot tasks, corresponding to four different settings respectively, i.e., ResNet12 on \emph{miniImageNet}, Conv4 on \emph{miniImageNet}, ResNet12 on \emph{CIFAR-FS} and Conv4 on \emph{CIFAR-FS}. For each task, we use the support samples, the adjusted support samples and the query samples to learn a principal component analysis (PCA) model which projects the feature representations into 2-D space. Then we apply this learned PCA model to all samples and the result is shown in Figure \ref{change}. 

Interestingly, Inverse Attention Module pushes the support samples away from their clusters, which reflects the difference between the LSSVM base learner and the nearest neighbor method and indicates Inverse Attention Module has adapted to the LSSVM base learner. Besides, the feature representations from the backbone have a cluster structure and samples of the same class are close to each other.

\subsubsection{Iteration number for Pseudo Support Module}
As discussed above, Pseudo Support Module can be iterated multiple times, and the new pseudo support samples can be added each time. So we explore the impact of the iterations number $k$ on accuracy. Specifically, we examine the variation of the accuracy increment with the iteration number $k$ under four different settings as above. We calculate the increment of the average accuracy, and the result is shown in Figure \ref{iteration}.

We can make the following observations: 1) as the iterations number increases, Pseudo Support Module can continuously improve the classification accuracy; 2) the increment of the 5-way 1-shot tasks is significantly higher than that of the 5-way 5-shot tasks and the reason may be that the information in the support set of the 1-shot tasks is too scarce to get more significant improvement using the same information increment; 3) this process cannot keep increasing the accuracy, but has a performance limit.

\section{Conclusion}
In this work, we analyze three directions where we can further improve few-shot learning, i.e., features suitable for comparison, the base learner suitable for low-data scenarios, and valuable information from the samples to classify, and make improvements in the last two directions. We first introduce multi-class least squares support vector machine as the base learner, which achieves better generalization and faster inference than existing ones. Then we propose two transductive modules that modify the support set using the query samples, i.e., adjusting the support samples and adding pseudo support samples. Experiments show that our transductive modules can significantly improve the few shot learning, especially for the difficult 1-shot setting. Note that the existing methods to make features more suitable for comparison, e.g., using self-supervised auxiliary training or using data augmentation (regional dropout), are compatible with our method and it can further improve performance when combining with them. For future work, we can make more explorations from these three directions.

\section*{Supplementary Material}
\subsubsection{Dataset} We evaluate our model on two standard few-shot learning benchmarks: \emph{miniImageNet} \cite{vinyals2016matching} and \emph{CIFAR-FS} \cite{bertinetto2018meta}. The \emph{miniImageNet} is the subset of ImageNet \cite{russakovsky2015imagenet} and includes a total of 100 classes with 600 images per class, and each image is of size $84\times84$. Following the setup provided by \cite{ravi2016optimization}, we use 64 classes as meta-training set, 16 and 20 classes as meta-validation set and meta-testing set respectively. The \emph{CIFAR-FS} consists of all 100 classes from CIFAR-100 \cite{krizhevsky2010cifar} and each class contains 600 images of size $32\times32$. All classes are split into 64, 16 and 20 for meta-training, meta-validation and meta-testing as in \cite{lee2019meta,yang2020dpgn}.
 
\subsubsection{Training scheme}
During meta-training, we perform data augmentation, such as horizontal flip, random crop and color (brightness, contrast and saturation) jitter, as in \cite{gidaris2018dynamic,qiao2018few,lee2019meta,yang2020dpgn}. We use SGD with Nesterov momentum of 0.9 and weight decay of 0.0005. The model is meta-trained for 60 epochs and each epoch consists of 1000 batches with each batch consisting of 8 episodes. Without pretraining initialization, all learning rate is initially set to 0.1, and with pretraining initialization, the initial learning rate of the backbone is set to 0.005 for Conv4 and 0.0005 for ResNet12. The learning rate of the Inverse Attention Module is initialized to 0.005 with ResNet12 as the backbone and 0.01 with Conv4 as the backbone so as to coordinate with different backbones. We drop the learning rate by the factor 0.06, 0.2, 0.2 at epoch 20, 40 and 50 respectively. 

As suggested in \cite{liu2018learning,lee2019meta}, we apply "Higher Shot" for meta-training which means keeping meta-training shot higher than meta-testing shot. Specifically, we set training shot to 5 for \emph{CIFAR-FS}, 5 for \emph{miniImageNet} 1-shot and 15 for \emph{miniImageNet} 5-shot. Each class contains 6 query samples during meta-training and 15 query samples during meta-testing. The optimal model is chosen on 5-way 5-shot tasks from the meta-validation set. 

The regularization parameter $\gamma$ of LSSVM is set to 0.1 for meta-training. And we use one-vs-all multi-class coding method and linear kernel function for all experiments. The reduction ratio $r$ of Inverse Attention Module is set to 8 for \emph{CIFAR-FS} and 16 for \emph{miniImageNet}. 

\bibliography{ref}
\end{document}